\documentclass{ecai} 

\usepackage{latexsym}
\usepackage{amssymb}
\usepackage{amsmath}
\usepackage{amsthm}
\usepackage{booktabs}
\usepackage{enumitem}
\usepackage{graphicx}
\usepackage{color}
\usepackage[skip=10pt]{caption}
\usepackage{multirow}

\newcommand{\BibTeX}{B\kern-.05em{\sc i\kern-.025em b}\kern-.08em\TeX}

\begin{document}


\begin{frontmatter}


\paperid{1304} 


\title{Enhancing Retrieval and Managing Retrieval: \\A Four-Module Synergy for Improved Quality and Efficiency in RAG Systems}


\author[A]{\fnms{Yunxiao}~\snm{Shi}}
\author[A]{\fnms{Xing}~\snm{Zi}}
\author[A]{\fnms{Zijing}~\snm{Shi}}
\author[A]{\fnms{Haimin}~\snm{Zhang}}
\author[A]{\fnms{Qiang}~\snm{Wu}}
\author[A]{\fnms{Min}~\snm{Xu}\thanks{Corresponding author with email: Min.Xu@uts.edu.au.}}


\address[A]{University of Technology Sydney, Broadway, Sydney, 2007, NSW, Australia.}


\begin{abstract}
Retrieval-augmented generation (RAG) techniques leverage the in-context learning capabilities of large language models (LLMs) to produce more accurate and relevant responses. Originating from the simple 'retrieve-then-read' approach, the RAG framework has evolved into a highly flexible and modular paradigm. A critical component, the Query Rewriter module, enhances knowledge retrieval by generating a search-friendly query. This method aligns input questions more closely with the knowledge base. Our research identifies opportunities to enhance the Query Rewriter module to \textbf{Query Rewriter+} by generating multiple queries to overcome the \textit{Information Plateaus} associated with a single query and by rewriting questions to eliminate \textit{Ambiguity}, thereby clarifying the underlying intent. We also find that current RAG systems exhibit issues with \textit{Irrelevant Knowledge}; to overcome this, we propose the \textbf{Knowledge Filter}. These two modules are both based on the instructional-tuned Gemma-2B model, which together enhance response quality. The final identified issue is \textit{Redundant Retrieval}; we introduce the \textbf{Memory Knowledge Reservoir} and the \textbf{Retriever Trigger} to solve this. The former supports the dynamic expansion of the RAG system's knowledge base in a parameter-free manner, while the latter optimizes the cost for accessing external knowledge, thereby improving resource utilization and response efficiency. These four RAG modules synergistically improve the response quality and efficiency of the RAG system. The effectiveness of these modules has been validated through experiments and ablation studies across six common QA datasets. The source code can be accessed at \url{https://github.com/Ancientshi/ERM4}.
\end{abstract}
\end{frontmatter}


\section{Introduction}

Large Language Models (LLMs) represent a significant leap in artificial intelligence, with breakthroughs in generalization and adaptability across diverse tasks \cite{zero_shot_llm, PaLM}. However, challenges such as hallucinations \cite{ReAct}, temporal misalignments \cite{temporal_adaptation}, context processing issues \cite{L_Eval}, and fine-tuning inefficiencies \cite{shortcut} have raised significant concerns about their reliability. In response, recent research has focused on enhancing LLMs' capabilities by integrating them with external knowledge sources through Retrieval-Augmented Generation (RAG) \cite{improve_by_retrieve, RAG, Atlas, DPR}. This approach significantly improves LLMs' ability to answer questions more accurately and contextually. 

The basic RAG system comprises a knowledge retrieval module and a read module, forming the retrieve-then-read pipeline \cite{RAG, DPR, Atlas}. However, this vanilla pipeline has low retrieval quality and produces unreliable answers. To transcend this, more advanced RAG modules have been developed and integrated into the basic pipeline. For example, the Query Rewriter module acts as a bridge between the input question and the retrieval module. Instead of directly using the original question as the query text, it generates a new query that better facilitates the retrieval of relevant information. This enhancement forms the Rewrite-Retrieve-Read pipeline \cite{Query_Rewriting, RETA_LLM}. Furthermore, models like RETA-LLM \cite{RETA_LLM} and RARR \cite{RARR} integrate a post-reading and fact-checking component to further solidify the reliability of responses. Additional auxiliary modules such as the query router \cite{query_router} and the resource ranker \footnote{https://txt.cohere.com/rerank/} \cite{LLMLingua} have also been proposed to be integrated into the RAG's framework to improve the practicality in complex application scenario. This integration of various modules into the RAG pipeline leading to the emergence of a modular RAG paradigm \cite{modular_rag_survey}, transforming the RAG framework into a highly flexible system.

\begin{figure*}[t]
  \centering
    \includegraphics[width=6.2 in]{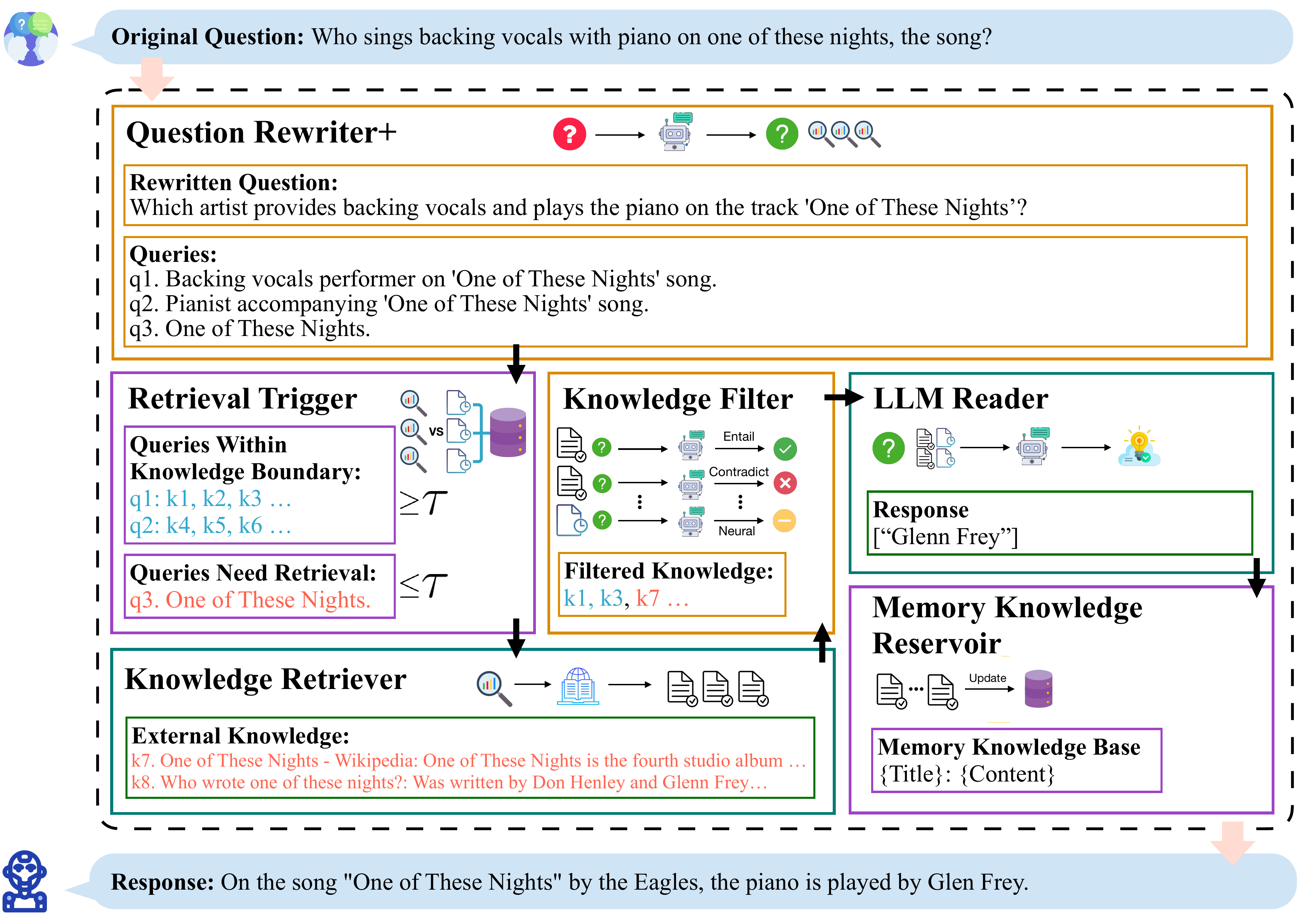}
   \caption{Flowchart depicting the integration of four modules into the basic Retrieve-then-Read (green) pipeline to enhance quality (orange) and efficiency (purple). Blue text represents cached knowledge retrieved from the Memory Knowledge Reservoir, while orange text indicates externally retrieved knowledge.\\\\}
    \label{fig: framework}
\end{figure*}

\vspace{40pt} 

Despite significant advancements, several unresolved deficiencies persist in practical applications. In the Query Rewriter module, the reliance on generating a single query for retrieval leads to (1) \textit{Information Plateau}, as this unidirectional search method limits the scope of retrievable information. Besides, the frequent misalignment between the input question and the underlying inquiry intent often exacerbated by (2) \textit{ambiguous phrasing}, significantly impedes the LLM's accurately interpreting users' demand. Furthermore, while the Query Rewriter can facilitate the retrieval of relevant information, it cannot guarantee the accuracy of the retrieved information. The extensive retrieval process may also acquire (3) \textit{irrelevant knowledge}, which detracts from the response quality by introducing noise into the context. At last, we have identified a phenomenon of (4) \textit{redundant retrieval}, where users pose questions similar to previous inquiries, causing the RAG system to fetch the same external information repeatedly. This redundancy severely compromises the efficiency of the RAG system.

These deficiencies underscore the potential for enhancing the accuracy and efficiency of existing RAG framework, thereby guiding our investigative efforts to address these issues. We decompose the Query Rewriter module’s functionality into two subtasks, resulting in the upgraded module \textbf{Query Rewriter+}. The first subtask involves crafting nuanced, multi-faceted queries for improving the comprehensive search of relevant information. The second subtask is rewriting the input text into a more intention-clear question. Given that both subtasks can be conceptualized as text-to-text task, we have designed them to be executed concurrently for high efficiency. This is achieved through parameter-efficient fine-tuning of the Gemma-2B model, serving as the backbone of the Query Rewriter+. The model is trained on a meticulously constructed supervised dataset, enabling it to efficiently generate both appropriate queries and rewritten question from an original input text. To address the issue of retrieving irrelevant knowledge, we introduce the \textbf{Knowledge Filter} module, which performs the Natural Language Inference (NLI) task to sift through retrieved information and assess its relevance. This NLI model is also based on Gemma-2B and fine-tuned on a carefully designed dataset. The synergistic use of these two modules demonstrably enhances the accuracy of RAG responses across various datasets.

To address the issue of redundant retrieval, we introduce the \textbf{Memory Knowledge Reservoir}, a non-parametric module that enhances the RAG system's knowledge base through a caching mechanism. This module facilitates rapid information retrieval for recurring queries, thereby eliminating redundant external knowledge search. To avoid the situation that the retrieved cached information is insufficient, we design the \textbf{Retrieval Trigger}, which employs a simple and effective calibration-based approach, to determine whether to engage external knowledge retrieval.

These four modular advancements work synergistically to enhance the RAG framework, significantly improving the accuracy and efficiency of responses. The integration and functionality of our proposed modules within the RAG system are illustrated in Figure~\ref{fig: framework}. In summary, our main contributions are as follows:

\begin{itemize}
\item We highlight the significance of clarifying input text and generating various queries, which informs the propose of Query Rewriter+. Furthermore, Knowledge Filter module is introduced to mitigate irrelevant knowledge issue. The synergistic operation of these two modules consistently enhances the response accuracy of RAG systems.
\item We pinpoint the problem of redundant retrieval within the current RAG system. To address this efficiency concern, we introduce the Memory Knowledge Reservoir and the Retrieval Trigger module.
\item Empirical evaluations and ablation studies across six QA datasets demonstrate the superior response accuracy and enhanced efficiency of our methods. Overall, our approach yields a 5\%-10\% increase in correctly hitting the target answers compared to direct inquiry. For historically similar questions, our method reduces the response time by 46\% without compromising the response quality.
\end{itemize}

\section{Motivation}
\label{sec:pls}
In this section, we empirically investigate several preliminary studies to highlight the limitations of current RAG systems in Open-Domain QA tasks.
\begin{itemize}
    \item \textbf{PS 1:} We investigate the maximum amount of relevant information that can be retrieved by converting input text into a single search-friendly query.
    \item \textbf{PS 2:} We examine whether using multiple queries that focus on different detailed semantic aspects can retrieve more relevant information than a single query.
    \item \textbf{PS 3:} We analyze how the proportion of irrelevant information changes as the volume of retrieved data increases.
    \item \textbf{PS 4:} We assess whether clarifying the input question is unnecessary for LLMs with strong semantic understanding capabilities.
\end{itemize}

\begin{figure}[t]
  \centering
    \includegraphics[width=3.3 in]{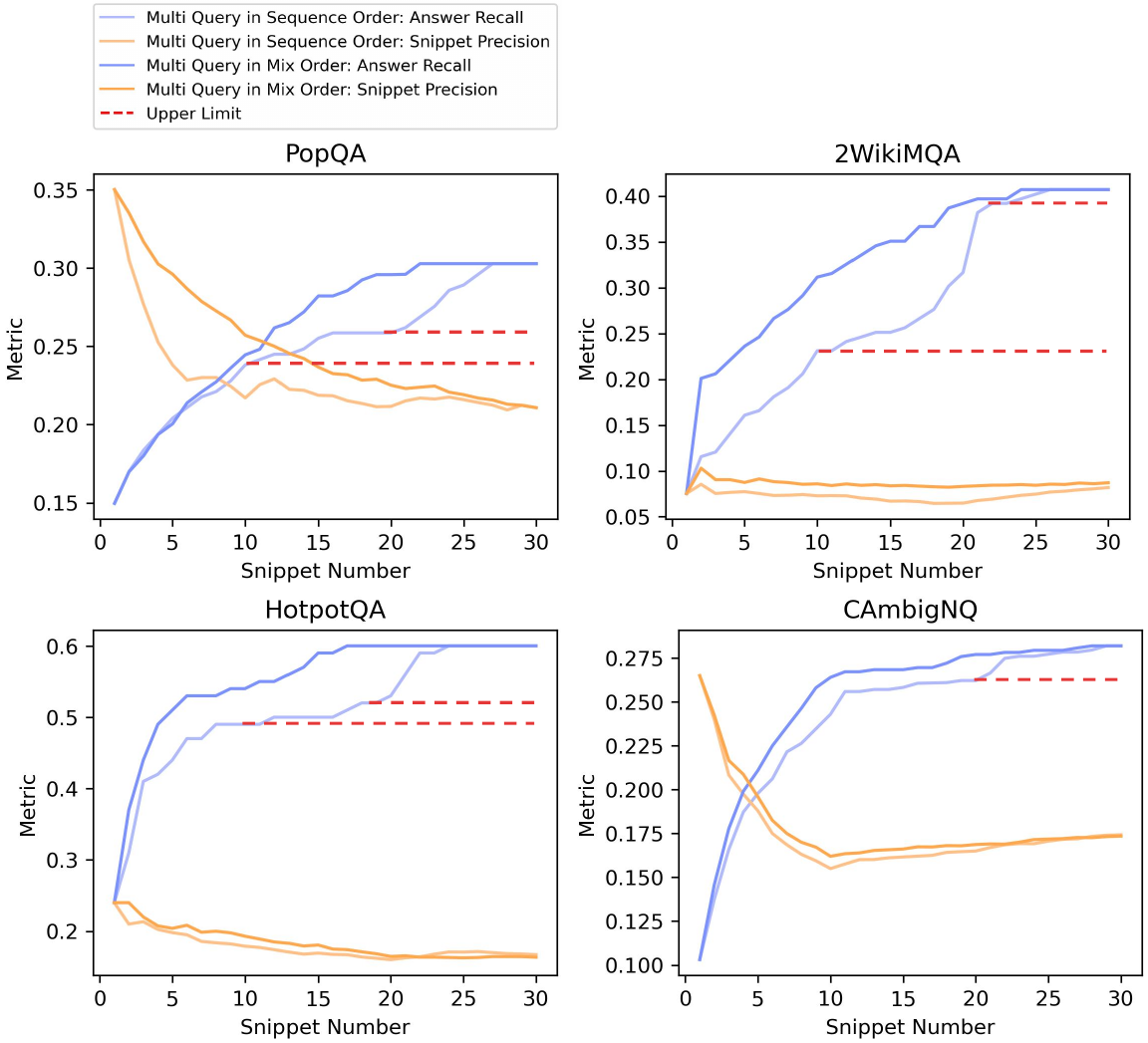}

   \caption{PS1 \& PS2. Analysis of Information Plateaus in Retrieving Knowledge and Overcoming the Bottlenecks with New Queries.\\\\} 
    \label{fig: pls1}
\end{figure}

\noindent \textbf{Experimental Settings.}  The experiment is conducted using a randomly selected subset of 50 questions from each of the following datasets: PopQA \cite{when_not_trust_llm}, 2WikiMQA \cite{2wikimqa}, HotpotQA \cite{hotpotqa}, and CAmbigNQ \cite{CAmbigNQ}. We employ the Rewrite-Retrieve-Read RAG pipeline, and the rewriter follows the LLM-based method, specifically utilizes GPT-3.5-turbo-0613, more details are described in previous study \cite{Query_Rewriting}. The rewriter module is designed to generate one to three variable-length queries for each question, depending on the question's complexity. For retrieval, we use Bing Search V7 to identify the top 10 most relevant webpage snippets for each query. These snippets encapsulate the most query-related content from each webpage. For each question, a collection of retrieved snippets serves as the external knowledge to facilitate the LLM’s in-context learning for generating response. We create two ways to present the snippets: \textit{Sequential Order}, with snippets for each query following one another, and  \textit{Mix Order}, with top snippets evenly sampled from different queries. We increase the number of snippets to see how retrieval performance changes in two different snippet arrangements. The evaluation metrics are Answer Recall, which is the ratio of answer items found in the external knowledge to the total number of answer items, and Snippet Precision, which is the ratio of snippets containing any answer item to the total number of snippets used. These metrics provide a quantitative measure of retrieval performance. The experimental results are illustrated in Figure~\ref{fig: pls1}.

\begin{figure}[t]
  \centering
    \includegraphics[width=3.3 in]{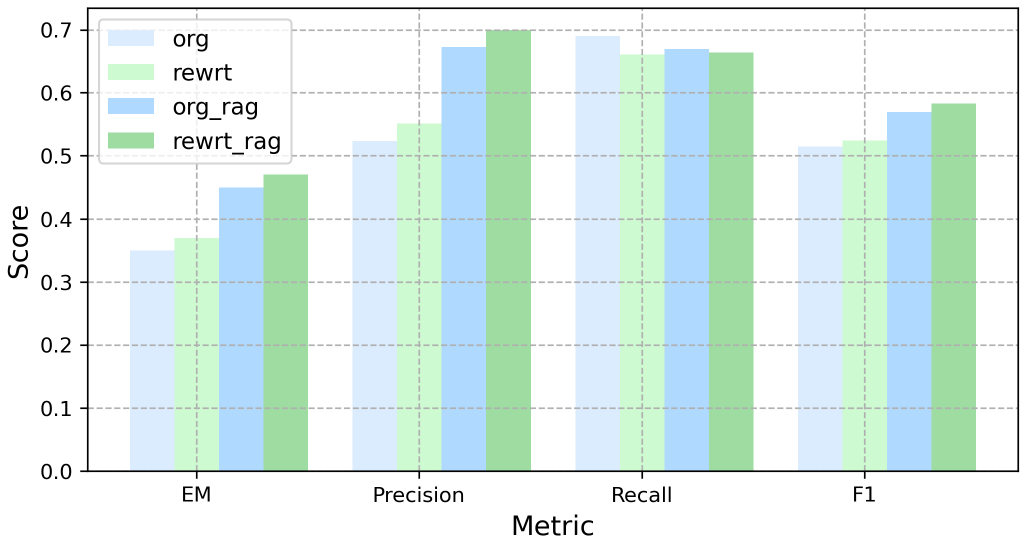}

   \caption{PS3. Evaluating the Impact of Rewriting Question for Q\&A.\\\\}
    \label{fig: pls2}
\end{figure}

We conduct another experiment using 50 questions from the CAmbigNQ dataset. We obtain responses by directly inputting the original questions into the LLM, labeled as \textit{org}. Another set of responses, labeled as \textit{rewrt}, is generated by inputting rewritten questions into the LLM. The accuracy of these two sets of responses is quantified using the metrics: EM (Exact Match), Precision, Recall, and F1 Score. Additionally, we also use retrieval-enhanced method to generate answers again. These responses are labeled as \textit{org\_rag} for original questions and \textit{rewrt\_rag} for rewritten questions. A comparative analysis of the answer accuracy is conducted in the same manner. These results are illustrated in Figure~\ref{fig: pls2}.

\vspace{40pt} 

\vspace{40pt} 

\noindent \textbf{PS1: The Information Plateaus of Single Query.}
From the analysis of Answer Recall in sequential order, it is observed that as the number of snippets increases, the Answer Recall metric improves, but they commonly plateauing before reaching 10 and 20 snippets (red dashed line). This phenomenon indicates there exist an upper limit to the retrievable useful information of a single query. This suggests there is a threshold beyond which additional information retrieved by the same query does not contribute too much to better retrieval quality.

\noindent \textbf{PS2: The Effect of Using Multiple Queries.} The analysis of Answer Recall in Sequence Order indicates that introducing fresh snippets from new queries effectively mitigates plateauing in Answer Recall (light purple solid line). Additionally, the Answer Recall in Mixed Order consistently outperforms that in Sequence Order at the same snippet number (purple solid line), particularly when the number has not yet reached the maximum retrievable information limit (30 snippets). This underscores the significance of multiple queries in enhancing retrieval quality.

\noindent \textbf{PS3: The Low Relevance of Retrieved Information.} As shown in Figure~\ref{fig: pls1}, Snippet Precision notably decreases as the number of snippets increases, eventually stabilizing. This suggests a significant presence of retrieved external knowledge snippets that do not contain relevant answer information.

\noindent \textbf{PS4: The Effect of Rewriting Questions.}
Figure~\ref{fig: pls2} presents a bar graph comparing various metrics for original and rewritten questions in the CAmbigNQ dataset. Rewriting questions improves Exact Match (EM), Precision, and F1 Score—whether or not retrieval augmentation technique is used. However, the recall decreases with rewritten questions. This occurs because the CAmbigNQ dataset labels include all possible answers, and LLMs tend to provide all possible responses to vague questions. The Rewritten questions are more well-intended, prompting LLMs to generate specific answers.

\noindent \textbf{Summary.}
Based on above results from a series of experiments on various datasets, our findings reveal that: (1) single query have an inherent upper limit of retrievable relevant information; (2) employing multiple queries that focus on different semantic aspects can surpass the information plateau, enhancing both the precision and recall of information retrieval; (3) The phenomenon of irrelevant knowledge is pervasive in RAG and becomes more pronounced with larger volumes of retrieved external information; and (4) rewriting ambiguous questions into intent-specific questions improves the precision of responses.


\section{Methodology}
\subsection{Question Rewriter+}
\label{sec: query_rewriter_plus}
The design of the Question Rewriter+ module encompasses two primary functions: (1) enhancing the original question semantically into a rewritten question, and (2) generating multiple search-friendly queries. Formally, the Question Rewriter+ is denoted as $G_\theta(\cdot)$, which takes an original question $p$ as input:

$$
G_\theta(p) \rightarrow (s, Q)
$$

where $s$ represents the rewritten question and $Q=\{q_1, q_2, \cdots, q_{|Q|}\}$ is the set of generated queries. A basic implementation of $G_\theta(\cdot)$ can adopt a prompt-based strategy, utilizing task descriptions, the original question, and exemplars to prompt black-box large language models. This approach capitalizes on the model's in-context learning capabilities, often yields effectiveness of question rewriting and query generation. Nevertheless, the effectiveness of this methodology is highly dependent on the meticulous construction of prompts tailored to specific domain datasets, which limits its general utility. Besides, the generated $s$ and $Q$ may be of low quality, failing to enhance RAG performance.  

To address these limitations, we propose a more general and task-specific approach. This involves parameter-efficient LoRA \cite{lora} fine-tuning of the Gemma-2B model for $G_\theta(\cdot)$, utilizing a high-quality dataset that is semi-automatically constructed through LLMs’ generation and human's quality validation. This dataset comprises instances $(p, s, Q)$, each rigorously validated to ensure that responses derived from $s$ are more accurate in hitting the labeled answer compared to those obtained by directly asking the LLM with $p$. Additionally, we manually verify the quality of generated queries $Q$ to ensure the reliability. The prompt template for generating $(s, Q)$ is as follows: 

\setlength{\fboxrule}{0.7pt}
\noindent  
\fbox{%
  \begin{minipage}{3.3 in}

\textbf{[Instruction]}: Your task is to transform a potentially colloquial or jargon-heavy [Original Question] into a semantically enhanced Rewritten Question with a clear intention. Additionally, generating several search-friendly Queries that can help find relevant information for answering the question. You can consider the provided [Examples] and response following the [Format].

\textbf{[Original Question]}:
\textit{\{User's original question is here.\}}

\textbf{[Examples]}:
\textit{\{The examples should be specially tailored for different datasets.\}}

\textbf{[Format]}:
\textit{\{The generated Rewritten Question is here\}**\{query1\}**\{query2\}**\{query3\}...}
  \end{minipage}%
}

\subsection{Knowledge Filter}
The accuracy of responses generated by LLMs can be significantly compromised by noisy retrieved contexts \cite{LLM_filter}. To mitigate this, we introduce the Knowledge Filter module, designed to enhance response accuracy and robustness. This module utilizes LLMs to filter out irrelevant knowledge. Rather than directly querying an LLM to identify noise, we incorporate a Natural Language Inference (NLI) framework \cite{NLI} for this purpose. Specifically, for a rewritten question $s$ and retrieved knowledge $k$, the NLI task evaluates whether the knowledge (as the premise) contains reliable answers, or useful information aiding the response to the question (as the hypothesis). This results in a judgment $j$ categorized as entailment, contradiction, or neutral. The operation of the Knowledge Filter can be mathematically represented as:
$$
F_\theta(s,k) \rightarrow j \in \{\text{entailment}, \text{contradiction}, \text{neutral}\}
$$

Knowledge is retained if the NLI result is classified as entailment. We can adjust the strength of the hypothesis based on the specific dataset. For single-hop questions, a stronger hypothesis can be set, requiring the knowledge to contain direct and explicit answer information. Conversely, for more complex multi-hop questions, we can set a weaker hypothesis, only requiring the knowledge to include information that possibly aids in answering the question. When valid knowledge is unavailable, a back-off strategy is invoked, where LLMs generate responses without the aid of external knowledge augmentation. The Knowledge Filter also employs the LoRA fine-tuning method \cite{lora} on the Gemma-2B model, offering enhanced applicability and adaptability compared to prompt-based approaches.

The NLI training dataset is constructed semi-automatically using the similar method we described in Section~\ref{sec: query_rewriter_plus}. We provide task instruction, rewritten question $s$, along with knowledge context $k$ as prompt to GPT-4, which then generated a brief explanation $e$ and a classification result $j$, resulting in the data instance $((s, k, (e,j))$. The prompt template is as follows:

\setlength{\fboxrule}{0.7pt}
\noindent  
\fbox{%
  \begin{minipage}{3.3 in}

\textbf{[Instruction]}: Your task is to solve the NLI problem: given the
premise in [Knowledge] and the hypothesis that "The [Knowledge] contains reliable answers aiding the response to
[Question]". You should classify the response as entailment, contradiction, or
neutral.

\textbf{[Question]}:
\textit{\{Question is here.\}}

\textbf{[Knowledge]}:
\textit{\{The judging knowledge is here.\}}

\textbf{[Format]}:
\textit{\{The explanation.\}**\{The NLI result.\}}

  \end{minipage}%
}

Considering that LLMs are primarily designed for text regression rather than classification, using only $j$ as a label for instructional tuning for Gemma-2B would prevent the LLM from accurately performing classification tasks in a generative manner. Therefore, we also incorporate the concise explanation $e$ as part of the label, in addition to the NLI classification result $j$.

\subsection{Memory Knowledge Reservoir}
We present a Memory Knowledge Reservoir module designed to cache the retrieved knowledge. The knowledge is structured as title-content pairs, where the title serves as a brief summary and the content offers detailed context. The Memory Knowledge Reservoir updates by adding new title-content pairs and replacing older entries with newer ones for the same titles. Mathematically, the Memory Knowledge Reservoir can be represented as a set $ K = \{k_1, k_2, \ldots, k_{|K|}\} $, where each $ k_i $ is a title-content pair.

\subsection{Retrieval Trigger}

This module assesses when to engage external knowledge retrieval. A calibration-based method is utilized, wherein the popularity serves as a metric to estimate a RAG system's proficiency with the related knowledge.

$ K = \{k_1, k_2, \ldots, k_{|K|}\} $ is the set of knowledge in the Memory Knowledge Reservoir, and $ q_i \in Q $ is a generated query. The cosine similarity between query $ q_i $ and a knowledge instance $ k_j \in K $ is denoted by $ S(q_i, title(k_j)) $. The popularity of query $ q_i $, denoted by $ \text{Pop}(q_i) $, is defined as:
$$\text{Pop}(q_i) = \left| \{k_j \in K \mid S(q_i, title(k_j)) \geq \tau \} \right|$$
where  $ \tau $  is a similarity threshold, $ | \cdot | $ indicates the cardinality of the set. The boundary conditions for a query being within or outside the knowledge of the RAG system are established using a popularity threshold $ \theta $. A query $ q_i $ is considered to be within the knowledge boundary if:
$$ \text{Pop}(q_i) \geq \theta $$

Conversely, a query $ q_i $ is outside the knowledge boundary if:
$$ \text{Pop}(q_i) < \theta $$


\section{Experiments}
\label{sec: exp}
\subsection{Modular Setting}
\noindent \textbf{Fine-tuning Gemma-2B}
We follow the Alpaca's training method\footnote{https://github.com/tloen/alpaca-lora}, employing the LoRa\cite{lora} method to instruction-tune the pre-trained Gemma-2B model\footnote{https://huggingface.co/google/gemma-2b} for the Question Rewriter+ and Knowledge Filter modules. We set the learning rate to 1e-4, batch size to 8, and epochs to 6. We set the rank of the LoRa low-rank matrix to 8, and the scaling factor, alpha to 16. Additionally, we utilize the 4-bit quantization method with NF4 quantization type \cite{qlora}. The training and inference process are all conducted on a single Nvidia Quadro RTX 6000.

\noindent \textbf{Knowledge Retriever} 
We utilize the Bing Search Engine v7 as the information retrieval method. For each query $ q $, we select the top-$ n $ items from the search results, and each item is regarded as a knowledge instance. We utilize the snippet of a search item as the content of a knowledge instance. The hyperparameter $ n $ is predetermined at 10. 

\noindent \textbf{LLM Reader}
We primarily used GPT-3.5-turbo-0613 as the black-box LLM model for generating answers. The prompt structure includes task instruction, question, external knowledge, examples, and response format. The external knowledge section comprises up to 30 knowledge instances arranged in mixed order, as discussed in Section~\ref{sec:pls}.

\subsection{Task Setting}
We evaluate the efficacy of our proposed methodologies under open-domain QA task. This evaluation leverages three distinct open-domain QA datasets that do not require logical reasoning. These include: (i) The Natural Questions (NQ) dataset \cite{NQ}, which is a real-world queries compiled from search engines. (ii) PopQA \cite{when_not_trust_llm}, a dataset with a long-tail distribution, emphasizing less popular topics within Wikidata. (iii) AmbigNQ \cite{AmbigQA}, an enhanced version of NQ that transforms ambiguous questions into a set of discrete, yet closely related queries. Additionally, we incorporate two benchmark datasets that require logical reasoning: (iv) 2WIKIMQA \cite{2wikimqa} and (v) HotPotQA \cite{hotpotqa}. Due to the costs associated with API calls for LLMs and Bing Search, and following common practices\cite{demonstrate,LLM_filter,Query_Rewriting,meta_answer}, we test on a stratified sample of 300 questions from each dataset rather than the entire test dataset.

We assess performance using the F1 Score and Hit Rate metrics. Due to the discrepancy between the verbose outputs of LLMs and the concise format of the dataset answers, we chose not to utilize the Exact Match (EM) metric. Instead, we considered a response as correct if the text hit any item of the labeled answers.

\subsection{Baselines}
Rewrite-Retrieve-Read \cite{Query_Rewriting} represents the current state-of-the-art improvement on the basic Retrieve-then-Read RAG pipeline. Our approach enhances the existing RAG pipeline by augmenting the Query Rewriter to Query Rewriter+ and introducing a new Knowledge Filter module. To highlight the effectiveness of our method, we implemented the following configurations:

(i) Direct: Ask the LLM directly with the original question.

(ii) Rewriter-Retriever-Reader: Prior to retrieval, a Query Rewriter module is employed to generate a query that fetches external knowledge. The external knowledge, along with the original question, is used to prompt the response generation.

(iii) Rewriter+-Retriever-Reader: Prior to retrieval, the Enhanced Query Rewriter module is utilized, generating multiple queries to acquire external knowledge and clarify the original question. Responses are generated using both the rewritten question and all retrieved external knowledge.

(iv) Rewriter+-Retriever-Filter-Reader: Applied before retrieval, the Enhanced Query Rewriter module generates multiple queries and clarifies the original question. A Knowledge Filter is used to discard external knowledge unrelated to the rewritten question. The final response is then generated using the filtered external knowledge and the rewritten question.

Comparing the Rewriter+-Retriever-Reader setup with the Rewriter-Retriever-Reader setup validates the superiority of the proposed Question Rewriter+ module. Additionally, comparing the Rewriter+-Retriever-Filter-Reader setup with the Rewriter+-Retriever-Reader setup demonstrates the effectiveness of the 4-step RAG pipeline incorporating the Knowledge Filter.

\begin{table}[t]
\caption{Open-Domain Question Answering Performance.}
\label{tab: open-domain}
\centering
\begin{tabular}{l|lll} 
\hline
Dataset                   & Method                            & F1             & Hit Rate        \\ 
\hline
\multirow{4}{*}{CAmbigNQ} & Direct                            & 37.38          & 55.67           \\
                          & Rewriter-Retriever-Reader         & 39.65          & 57.67           \\
                          & Rewriter+-Retriever-Reader        & 41.58          & 59.00           \\
                          & Rewriter+-Retriever-Filter-Reader & \textbf{43.39} & \textbf{64.33}  \\ 
\hline
\multirow{4}{*}{NQ}       & Direct                            & 41.50          & 42.00           \\
                          & Rewriter-Retriever-Reader         & 48.70          & 46.00           \\
                          & Rewriter+-Retriever-Reader        & 51.43          & 50.33           \\
                          & Rewriter+-Retriever-Filter-Reader & \textbf{52.68} & \textbf{51.33}  \\ 
\hline
\multirow{4}{*}{PopQA}    & Direct                            & 35.24          & 42.33           \\
                          & Rewriter-Retriever-Reader         & 37.84          & 44.00           \\
                          & Rewriter+-Retriever-Reader        & 38.51          & 46.67           \\
                          & Rewriter+-Retriever-Filter-Reader & \textbf{41.77} & \textbf{51.33}  \\ 
\hline
\multirow{4}{*}{AmbigNQ}  & Direct                            & 45.21          & 46.67           \\
                          & Rewriter-Retriever-Reader         & 49.85          & 50.00           \\
                          & Rewriter+-Retriever-Reader        & 51.84          & 52.33           \\
                          & Rewriter+-Retriever-Filter-Reader & \textbf{53.47} & \textbf{55.67}  \\ 
\hline
\multirow{4}{*}{HotPot}   & Direct                            & 46.33          & 41.33           \\
                          & Rewriter-Retriever-Reader         & 48.24          & 44.00           \\
                          & Rewriter+-Retriever-Reader        & 53.67          & 45.33           \\
                          & Rewriter+-Retriever-Filter-Reader & \textbf{57.59} & \textbf{50.00}  \\ 
\hline
\multirow{4}{*}{2WikiMQA} & Direct                            & 41.85          & 42.33           \\
                          & Rewriter-Retriever-Reader         & 43.24          & 45.00           \\
                          & Rewriter+-Retriever-Reader        & 45.71          & 47.67           \\
                          & Rewriter+-Retriever-Filter-Reader & \textbf{46.83} & \textbf{49.33}  \\
\hline
\end{tabular}
\end{table}

\subsection{Results}
Experimental results are reported in Table~\ref{tab: open-domain}. The scores indicate that the Query Rewriter+ module outperforms the Query Rewriter module across all datasets, substantiating that multiple queries and clarified question are more effective for a RAG system correctly response to user's questions than single query and unrefined questions. Specifically, adding the Knowledge Filter module to the traditional 3-step RAG pipeline significantly improves performance. This indicates that merely adding external knowledge to the RAG system can be detrimental, especially for multi-hop questions. The Knowledge Filter module effectively eliminates noise and irrelevant content, enhancing the accuracy and robustness of the RAG system's responses.

\begin{table*}[t]
\caption{The Impact of the Similarity Threshold $\tau$ of the Retriever Trigger module in Response Efficiency and Quality.}
\label{tab: eff}
\centering
\begin{tabular}{llllll} 
\toprule
$\tau$   & Time Cost (s) & External Knowledge & Memory Knowledge & Irrelevant Knowledge & Hit Rate (\%)  \\ 
\hline
0.2 & 5.68          & 0.00               & 30.30            & 11.62                & 48.5           \\
0.4 & 5.89          & 0.00               & 19.58            & 3.52                 & 50.5           \\
0.6 & 3.97          & 4.39               & 11.57            & 1.53                 & 53.0           \\
0.8 & 7.17          & 14.33              & 1.72             & 0.86                 & 55.0           \\
1.0 & 7.45          & 15.00              & 0.00             & 0.79                 & 53.5           \\
\bottomrule
\end{tabular}
\end{table*}

\begin{table}[t]
\caption{Investigation of the Individual and Combined Effects of Question Rewriting and Knowledge Filtering on CAMbigNQ and PopQA}
\label{tab: ablation_study}
\centering
\begin{tabular}{l|ll|ll} 
\hline
\multirow{2}{*}{Dataset}  & \multicolumn{2}{l|}{Setting}                     & \multicolumn{1}{l|}{\multirow{2}{*}{F1}} & \multirow{2}{*}{Hit Rate}  \\ 
\cline{2-3}
                          & \multicolumn{1}{l|}{Question} & Knowledge        & \multicolumn{1}{l|}{}                    &                            \\ 
\hline
\multirow{6}{*}{CAmbigNQ} & Original                      & \textbackslash{} & 37.38                                    & 55.67                      \\
                          & Rewritten                     & \textbackslash{} & 38.45                                    & 57.67                      \\
                          & Original                      & All              & 38.24                                    & 54.00                      \\
                          & Rewritten                     & All              & 41.58                                    & 58.00                      \\
                          & Original                      & Filtered         & 39.62                                    & 59.67                      \\
                          & Rewritten                     & Filtered         & 43.39                                    & 64.33                      \\ 
\hline
\multirow{6}{*}{PopQA}    & Original                      & \textbackslash{} & 35.24                                    & 42.33                      \\
                          & Rewritten                     & \textbackslash{} & 36.73                                    & 42.67                      \\
                          & Original                      & All              & 38.14                                    & 44.33                      \\
                          & Rewritten                     & All              & 38.51                                    & 46.67                      \\
                          & Original                      & Filtered         & 39.79                                    & 47.33                      \\
                          & Rewritten                     & Filtered         & 41.77                                    & 51.33                      \\
\hline
\end{tabular}
\end{table}

\section{Ablation Studies}
In this section, we analyze the individual and combined effects of question rewriting and knowledge filtering. The results presented in Table~\ref{tab: ablation_study} indicate that the question rewriting process consistently improves answer accuracy across the setups of direct generation, retrieval-augmented generation using all knowledge, and retrieval-augmented generation using filtered knowledge.

The results also shows that using all (unfiltered) external knowledge for retrieval-augmented generation can sometimes lead to marginal improvements or even decreased performance. For instance, on the CAmbigNQ dataset, when LLMs are asked with rewritten questions, introducing all external knowledge only raise the hit rate from 57.67\% to 58\%. Besides, when LLM are queried with the original questions, introducing all external knowledge makes the hit rate decreased from 55.67\% to 54.00\%. On the other hand, we observe that filtering knowledge can significantly boost the response accuracy of the RAG system, whether asking with the original question or rewritten question. 

Assessing the synergistic effect of two modules, we find that while each module individually improves response accuracy, the effect is sometimes modest. However, their combined yields a significant enhancement. For instance, on the CAmbigNQ dataset, the individual application of each module resulted in a maximum of 2\% more correctly answered questions, whereas their combined application led to a 7\% increase in correctly answered questions. A similar phenomenon can also been observed on the PopQA dataset.

\section{Efficiency Improvement Investigation}
In this section, we explore how efficiently our proposed method reduces redundant retrieval when answering recurring questions with historically similar semantics. We also examine the hyperparameter $\tau$ to balance efficiency and response accuracy. The experimental procedure is as follows:

Initially, we randomly selected 100 questions from the AmbigNQ dataset to generate responses using our proposed method. Unlike previous sections, we set the parameter $n$ in the Knowledge Retriever module to 5. Instead of utilizing webpage snippets as the content of knowledge instances, we visited the searched URLs and read the entire webpage text, filtering out irrelevant information using the BM25 algorithm. After the response finished, the webpage content was then cached in the Memory Knowledge Reservoir. Subsequently, we selected an additional 200 questions from AmbigNQ that are semantically similar to the previously solved questions. These questions were answered with the support of the Memory Knowledge Reservoir and the Retrieval Trigger module, with the popularity threshold $\theta$ set as 3.

We design several metrics to evaluate the resource cost of per-question, include the average time spent in the RAG pipeline (Time Cost), the average number of external knowledge instances (External Knowledge), the average number of memory knowledge instances (Memory Knowledge), the average number of knowledge instances filtered out (Irrelevant Knowledge), and the performance metric Hit Rate. These metrics are recorded during the question-answering process.

The analysis of the trade-off between response quality and efficiency for answering historically similar questions across different $\tau$ settings is presented in Table~\ref{tab: eff}. A significant finding is that the Time Cost metric reaches minimum when setting the similarity threshold $\tau=0.6$. This is accompanied by the External Knowledge metric being very small, approximately 4.39, which is roughly equivalent to one query search. This suggests that this configuration predominantly leverages memory knowledge rather than external sources for generating responses, thereby enhancing response efficiency. Remarkably, at the $\tau=0.6$ setup, the quality of the responses is not heavily affected and remains very close to that achieved by relying entirely on external knowledge at $\tau=1.0$. This suggests that deploying the Memory Knowledge module can achieve a significant reduction in response time—by approximately 46\%—without substantially compromising the quality of the answers. Furthermore, adjusting the threshold to 0.8 enhances the response quality beyond that at $\tau=1.0$, underscoring that leveraging highly relevant historical experience can generate responses with superior quality.

\section{Case Study}
\begin{figure}[t]
  \centering
    \includegraphics[width=3.3 in]{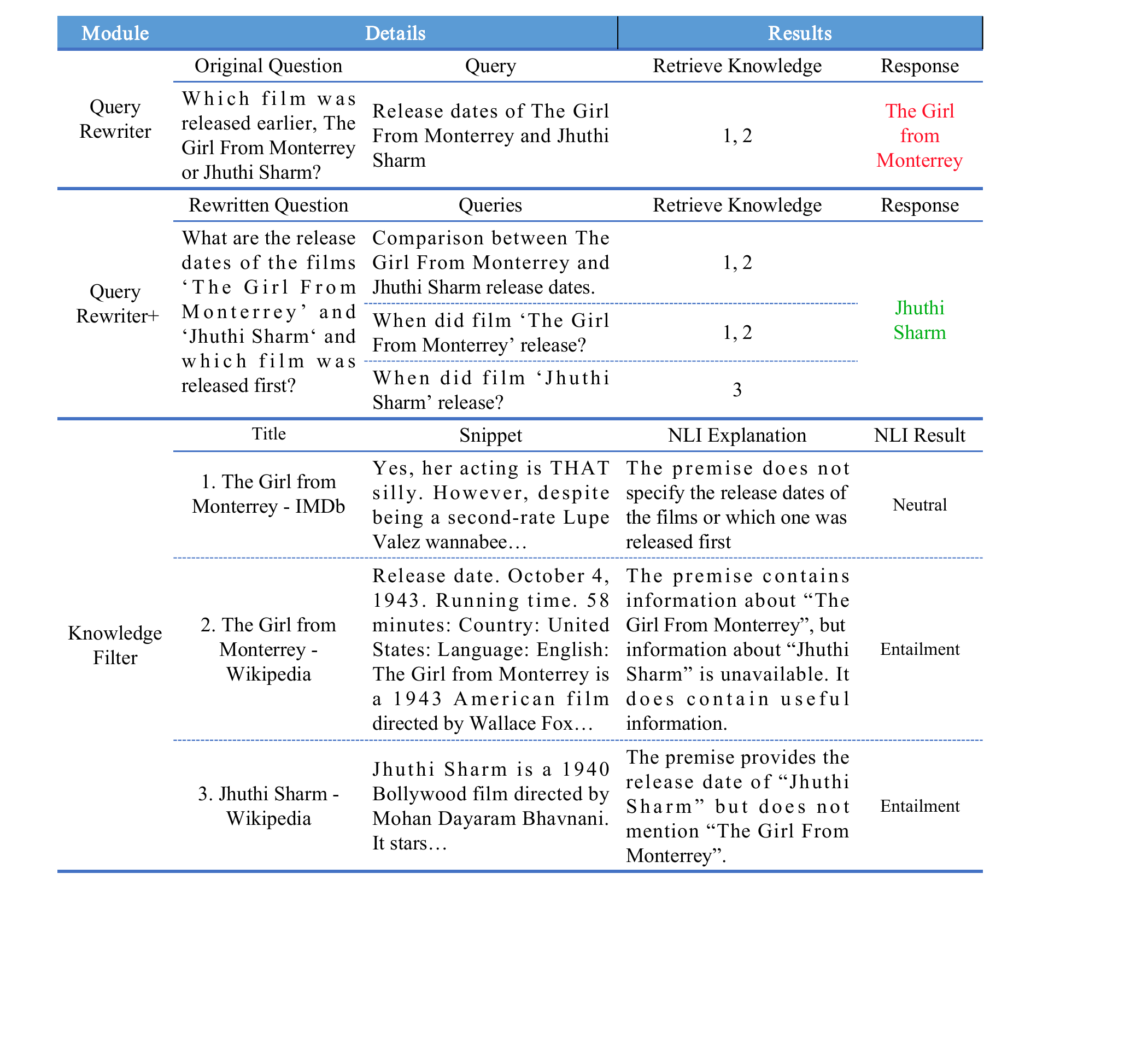}
   \caption{Examples for answering question "Which film was released earlier, The Girl From Monterrey or Jhuthi Sharm?" in the 2WIKIMQA dataset.\\\\}
    \label{fig: case_study}
\end{figure}

To intuitively demonstrate how the Query Rewriter+ Module enhances the original question and generates multiple queries, as compared to traditional Query Rewriter Modules, we present a question examples from 2WikiMQA dataset in Figure~\ref{fig: case_study}. It can be observed that the Query Rewriter+ Module semantically enhances the original question and, unlike the Query Rewriter which generates only one query, it produces three distinct queries, each focusing on different semantic aspects. The Query Rewriter+ module can retrieve external knowledge sources 1, 2, and 3, whereas the Query Rewriter module only retrieves sources 1 and 2, showcasing its advantage in improving knowledge recall. The Knowledge Filter module subsequently ensures the precision of the external knowledge by filtering out irrelevant knowledge instances (neutral, contradict) and retaining only those that provide valuable information for answering the question (entailment).

\vspace{40pt} 

\section{Related Work}
\subsection{Retrieval Augmented Generation}
Retrieval Augmented Generation (RAG) \cite{RAG} leverages a retriever that provides substantial external information to enhance the output of Large Language Models (LLMs). This strategy utilizes knowledge in a parameter-free manner and circumvents the high training costs associated with LLMs' parameterized knowledge. Furthermore, it alleviates hallucination issues, significantly enhancing the factual accuracy and relevance of the generated content. The concept of RAG is rooted in the DrQA framework \cite{DrQA}, which marked the initial phase of integrating retrieval mechanisms with Language Models through heuristic retrievers like TF-IDF for sourcing evidence. Subsequently, RAG evolved with the introduction of Dense Passage Retrieval \cite{DPR} and REALM \cite{IC-RALM}. These methods utilize pre-trained transformers and are characterized by the joint optimization of retrieval and generation process. Recent advancements have extended RAG's capabilities by integrating Large Language Models (LLMs), with developments such as REPLUG \cite{REPLUG} and IC-RALM \cite{IC-RALM} demonstrating the potent generalization abilities of LLMs in zero-shot or few-shot scenarios. These models can follow complex instructions, understand retrieved information, and utilize limited demonstrations for generating high-quality responses.

\subsection{Modular RAG}
The core of RAG framework consists of retriever and reader modules. This retrieve-read pipeline has been enhanced, leading to the Modular RAG paradigm with various integrated modules. This section describes related modules in our work.

\textit{Rewriter:} The introduction of a Question Rewriter module \cite{Query_Rewriting} led to the development of a Rewrite-Retrieve-Read RAG pipeline. This module generates a query that bridges the gap between the input text and the knowledge base, facilitating the retrieval of relevant knowledge and enhancing response accuracy. Our empirical studies indicate that while a single query retrieves limited useful information, multiple queries significantly enhance the retrieval of answer keywords. This discovery has reinforced and motivated our efforts to improve the existing functionality and design of the Question Rewriter.

\textit{Clarification:} Represented by \cite{TOC}, this module generates clarification questions to ascertain user intent, thus refining vague questions to uncover the underlying inquiry intent. We have integrated the functionalities of the Rewriter and Clarification modules into a single unit, Query Rewriter+, employing a fine-tuned Gemma-2B model to perform both tasks generatively in one step, improving efficiency.

\textit{Post-Retrieval Process:} After information retrieval, presenting all data to a Large Language Model simultaneously may exceed the context window limit. The Re-Ranking module strategically relocates content based on relevance. Our preliminary study reveals that Large Language Models (LLMs) have evolved to handle extended contexts, accommodating all retrievable information until a bottleneck is reached. Consequently, we consider this post-retrieval process primarily as a de-noising task, rather than focusing on ranking.

\textit{Memory:} Modules in this category leverage historically similar question-answer records to enhance current problem-solving capabilities, reflecting an evolutionary learning process within the agent \cite{Expel}. Drawing on this concept, we employ a parameter-free caching mechanism to expand the knowledge boundaries of RAG-based question-answering, effectively improving response efficiency.

\textit{Retrieve Trigger:} Understanding the parameterized knowledge boundaries of LLMs is crucial for optimizing the timing of knowledge retrieval. Calibration-based judgment methods have proven both efficient and practical. However, our study explores a non-parametric knowledge domain within a continuously expanding RAG system. This is the first attempt to design a Retrieve Trigger specifically for such scenarios. Our exploration focuses on identifying appropriate thresholds that balance accuracy and efficiency.

Additional modules include Knowledge Retriever, LLM Reader, Fact Checking, Revising \cite{RETA_LLM, RARR}, and iterative RAG pipeline \cite{ITER-RETGEN} with further details available in \cite{modular_rag_survey, feng2023trends}.

\section{Conclusion}
In this paper, we present a four-module strategy to enhance RAG systems. The Query Rewriter+ module generates clearer questions for better intent understanding by LLMs and produces multiple, semantically distinct queries to find more relevant information. The Knowledge Filter refines retrieved information by eliminating irrelevant and noisy context, enhancing the precision and robustness of LLM-generated responses. The Memory Knowledge Reservoir and the Retrieval Trigger module optimize the use of historical data and dynamically manage external information retrieval needs, increasing system efficiency. Collectively, these advancements improve the accuracy and efficiency of the RAG system.



\begin{ack}
This work was sponsored by the \texttt{Australian Research Council under the Linkage Projects Grant LP210100129}, and by the program of \texttt{China Scholarships Council (No. 202308200014)}.
\end{ack}



\bibliography{mybibfile}

\end{document}